\documentclass{egpubl}
 
\JournalSubmission    
\usepackage[T1]{fontenc}
\usepackage{dfadobe}
\usepackage{colortbl}
\usepackage{algorithm}
\usepackage{algpseudocode}
\usepackage{wrapfig}
\usepackage{subcaption}
\usepackage{amsmath}
\usepackage{amssymb}
\usepackage[table]{xcolor}

\biberVersion
\BibtexOrBiblatex
\usepackage[backend=biber,bibstyle=EG,citestyle=alphabetic,backref=true]{biblatex} 
\electronicVersion
\PrintedOrElectronic

\ifpdf \usepackage[pdftex]{graphicx} \pdfcompresslevel=9
\else \usepackage[dvips]{graphicx} \fi

\usepackage{egweblnk} 

\newcommand{\vm}[1]{\textcolor{black}{#1}}
\newcommand{\diss}{\mathbf{s}}
\newcommand{\disS}{\mathbf{S}}
\newcommand{\conts}{\tilde{\mathbf{s}}}

\algdef{SE}[VARIABLES]{Variables}{EndVariables}
   {\algorithmicvariables}
   {\algorithmicend\ \algorithmicvariables}
\algnewcommand{\algorithmicvariables}{\textbf{global input variables}}

\algdef{SE}[VARIABLES]{IPVariables}{EndIPVariables}
   {\algorithmicipvariables}
   {\algorithmicend\ \algorithmicipvariables}
\algnewcommand{\algorithmicipvariables}{\textbf{interior point solver parameters}}

\definecolor{greycolor}{rgb}{0.6,0.6,0.6}
\definecolor{redcolor}{rgb}{0.8,0,0}
\definecolor{bluecolor}{rgb}{0,0,0.8}
\definecolor{greencolor}{rgb}{0,0.7,0}
\definecolor{browncolor}{rgb}{0.5,0.2,0.2}
\definecolor{greycolor}{rgb}{0.6,0.6,0.6}
\definecolor{purplecolor}{rgb}{0.6,0.0,0.6}
\allowdisplaybreaks

\title{Multi-Agent Path Planning with Asymmetric Interactions In Tight Spaces}

\author[V. Modi \& Y. Chhen \& A. Madan \& S. Sueda \& D.\,I.\,W. Levin]
{\parbox{\textwidth}{\centering V. Modi$^{1}$\orcid{ 0000-0002-9350-494X} Y. Chen $^{1}$\orcid{0000-0001-7547-9587},
 A. Madan $^{1}$\orcid{0000-0001-7547-9587},
 S. Sueda $^{2}$\orcid{0000-0003-4656-498X},
 and D.\,I.\,W. Levin $^{1}$\orcid{0000-0001-7079-1934}
        }
        \\
{\parbox{\textwidth}{\centering $^1$University of Toronto, Toronto, Canada\\
         $^2$Texas A\&M University, College Station, TX
       }
}
}

\begin{document}

\teaser{
 \includegraphics[width=\linewidth]{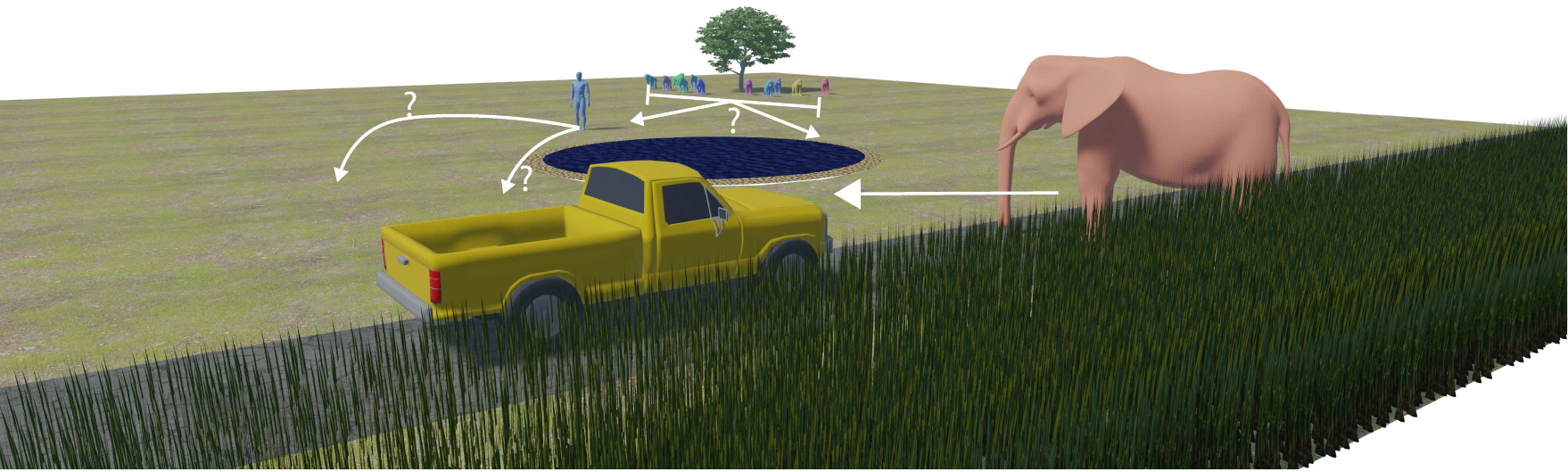}
 \centering
  \caption{Safari Escape}
\label{fig:teaser}
}

\maketitle
\begin{abstract}
By starting with the assumption that motion is fundamentally a decision making problem, we use \vm{the world-line concept from Special Relativity} as the inspiration for a novel multi-agent path planning method. We have identified a particular set of problems that have so far been overlooked by previous works. We present our solution for the global path planning problem for each agent and ensure smooth local collision avoidance for each pair of agents in the scene. We accomplish this by modeling the trajectories of the agents through 2D space and time as curves in 3D. Global path planning is solved using a modified Djikstra's algorithm to ensure that initial trajectories for agents do not intersect. We then solve for smooth local trajectories using a quasi-Newton interior point solver, providing the trajectory curves with a radius to turn them into rods. Subsequently, resolving collision of the rods ensures that no two agents are in the same spatial position at the same time. This space-time formulation allows us to simulate previously ignored phenomena such as highly asymmetric interactions in very constrained environments. It also provides a solution for scenes with unnaturally symmetric agent alignments without the need for jittering agent positions or velocities. 

\ccsdesc[500]{Computing Methodologies~ Applied computing}

\end{abstract}

\section{Introduction}
On a hot dry day in Kruger National Park, an empty truck idles on the side of a road. Sam, the driver of the truck, has wandered a hundred meters off the road in an attempt to take a picture of a tree with ten baboons. 
The baboons suddenly jump out of the tree and charge towards Sam. Sam panics and starts running back to the truck; however, coincidentally an elephant wandering in the area is on a path perpendicular to Sam’s, between him and the truck. 
How will Sam get back to the truck safely while outrunning the baboons and avoiding collision with the elephant?
What are the paths of the twelve agents in the scene: Sam, the elephant and the ten baboons? 

The problem above poses many challenges to a multi-agent path planning algorithm. 
First Sam must anticipate collisions ahead of time, in order to move quickly and efficiently to their truck. 
Second, the three groups of agents, Sam, the baboons and the elephant have dramatically different masses and behaviors.
For instance while Sam seeks to avoid all animals, the massive elephant is untroubled, and will stubbornly continue on its path. 
Finally, geographic features such as additional trees and ponds can lead to a highly constrained environment.  
Existing state-of-the-art crowd simulation methods struggle to compute anticipatory agent paths in constrained environments when asymmetric  interactions are involved (\autoref{table:crowds_vs_others}) making them ill-suited for application in planning problems such as the example given above. 

We propose a new multi-agent path planning algorithm well suited for these problems.
Our model directly optimizes the space-time trajectories of all agents  which allows for per-agent physical and psychological characteristics and smooth anticipatory trajectories. A novel, differentiable space-time  repulsive energy  ensures collision free trajectories. 
Using our approach, Sam arrives safely at the truck, escaping the baboons and avoiding the elephant. 

\begin{figure}[h]
\includegraphics[width=\columnwidth]{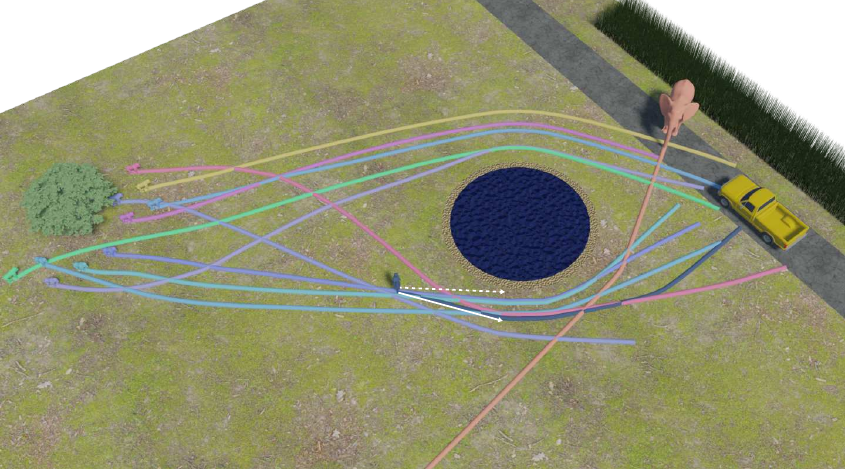}
\caption{ Sam needs to escape ,from the baboons while avoiding the elephant, whereas the elephant is unconcerned with the other agents in the scene.} 
\label{fig:ricky}
\end{figure}

\section{Related Work}
Successful multi-agent path planning requires an algorithm to both correctly model the behavior of independent agents as well as their interactions. 
A common approach is to apply a dynamics model based on Newton's second law of motion which is integrated over time to produce plausible agent trajectories.
Psychological and social characteristics of the group can be incorporated into the dynamics equations as social forces~\cite{pelechano2005crowd,socialforces,weiss2017position}. 
Intra-agent forces, such as those that handle collision avoidance are added through a variety of means, and we can partition the space of successful approaches based on locality of their models in both space and time. \vm{Approaches such as~\cite{guy2011simulating} incorporate psychological factors into an underlying dynamics model, but it is impossible to tell apart the psychological traits of the agents by simply observing the simulation. Our method makes the impact of agent characteristics on the trajectory very obvious, while providing a standalone local and global dynamics model for the scene.}

In local methods, agent decision making requires information only local in space and in time. Local collision avoidance methods~\cite{van2011reciprocal,Golas2013,wolinski2016warpdriver} compute collision response using local information in space and time (the planning horizon can be as small as a single time step). These methods struggle to generate smooth anticipatory collision responses. Implicit Crowds~\cite{karamouzas2017} attempts to overcome this difficulty by introducing a time-to-collision potential to the crowd dynamics. This gives agents richer space-time information on which to act; however this energy is effectively local, computed from the current state of the system (position and velocity). NH-TTC~\cite{nhttc} improves upon prior work by utilizing longer planning horizon, geometrically represented by curves in space. Intersection checks between these curves allow agents to react to collisions likely to occur in the near future. \vm{Similarly, vision based methods, such as~\cite{ondvrej2010synthetic,dutra2017gradient} provide an anticipatory collision avoidance model, but with extremely limited path planning capabilities and no guarantee of smooth agent motion. Data-driven aproaches such as \cite{charalambous2014pag} use an underlying state-action graph created via external data to generate trajectories. However, these trajectories are highly data-dependent, do not factor in environmental constraints, seem to operate in sparse crowds and must be used in conjunction with some other higher level global path planner.}  

An alternative approach to more local methods is to extend the collision response globally in space using a fluid like pressure solve~\cite{hughescontinuum,treuille2006continuum,narain2009aggregate}. However because these methods only consider the configuration of the system (position and velocity of each agent) at a single time, their ability to produce smooth anticipatory collision response, especially in a sparser setting, is reduced. \vm{Continuum Crowds stands out as one of the only methods that incorporates both a global planner through Djikstra's search and local collision avoidance through a fluid-like pressure solve. Another option is to handle local collision avoidance using RVO or Social Forces, as done by the data-driven method~\cite{takahashi2009spectral}. Optionally, one might use~\cite{hyun2013tilingmotion}, another data-driven method which uses a purely stochastic collision avoidance method. None of these approaches solve the problems of handling tight environmental constrains or asymmetric agent interactions. In fact, an adjacent field of research, that involves measuring the "correctness" of various crowd models (~\cite{guy2012statistical,wolinski2014parameter}) also fails to test for asymmetric agents and behavior in complex environments.}
\setlength{\columnsep}{1em}
\setlength{\intextsep}{0em} 
\begin{wrapfigure}{r}{0.5\columnwidth}
\centering
\includegraphics[width=0.5\columnwidth]{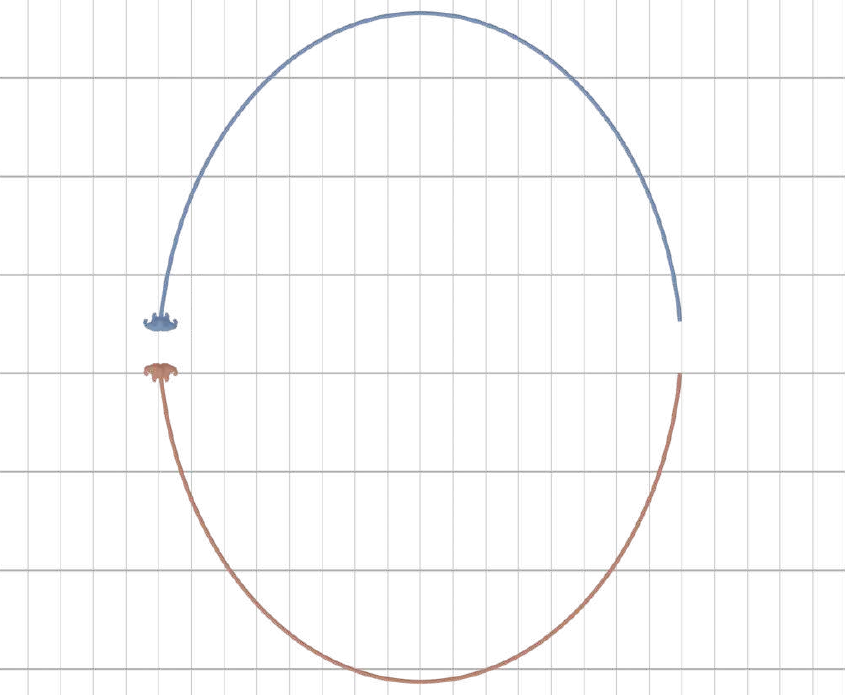}
\caption{The repulsive curves energy forces agents to be maximally far away from one another which is not an intuitive behavior.} 
\label{fig:repulsivecurves}
\end{wrapfigure}

\vm{Finally, while not strictly designed for multi-agent path planning, Repulsive Curves~\cite{repulsivecurves} can potentially be used for agent planning. While it yields impressive results in 3D curve untangling, when applied to multi-agent path planning it has several drawbacks. First, in unconstrained environments, Repulsive Curves will maximally repell agents away as shown in~\autoref{fig:repulsivecurves}. Secondly, like other previous methods, there is no straightforward way to model asymmetric agents. Lastly, Repulsive Curves uses random "jitter" to ensure that no trajectories initially overlap. From our own experiments, we have observed that this is not sufficient to guarantee non-overlapping initial trajectories. Meanwhile, our space-time Djikstra's approach is guaranteed to create initially non-intersecting trajectories, but is not sufficient for overcoming the other limitations of Repulsive Curves.}

In this paper we present a multi-agent path planner that computes its response \textit{globally in space and time}. 
Rather than formulating our multi-agent planner as an initial value problem, we instead take a space-time optimization approach. 

\vm{Space-time trajectory optimization~\cite{witkin1988spacetime} was initially applied to keyframe interpolation, with subsequent methods such as~\cite{popovic2000interactive} allowing manual editing of object trajectories and~\cite{shapiro2008interactive} allowing interactive motion correction and synthesis using graph search methods. The concept of using space-time graph search methods for collision avoidance is furthered in~\cite{SpacetimePlanning} for a single agent and~\cite{singh2011modular} for small groups. For larger situations, standard space-time approaches would be computationally prohibitive. Our method builds on these prior methods by expanding the look-ahead globally and ensuring smooth paths by collision resolution on the entire trajectories rather than using a limited space-time graph search approach.}

\begin{figure}[h]
\centering
  \begin{subfigure}[b]{0.49\columnwidth}
  \includegraphics[width=\columnwidth]{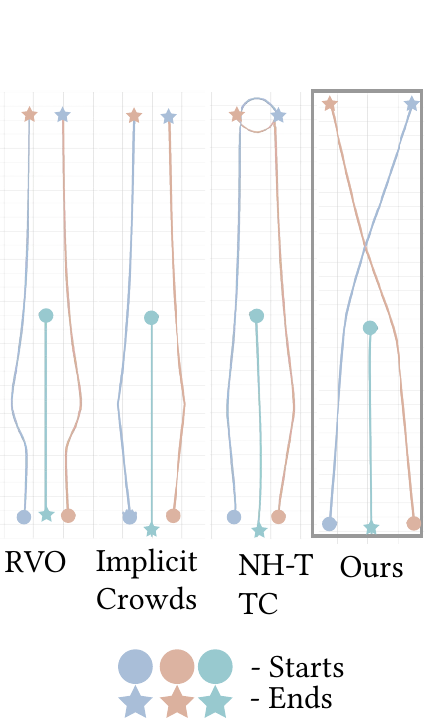}
  \caption{3 agents} 
  \label{fig:3personcomparisons}
  \end{subfigure}
  \begin{subfigure}[b]{0.49\columnwidth}
  \includegraphics[width=\columnwidth]{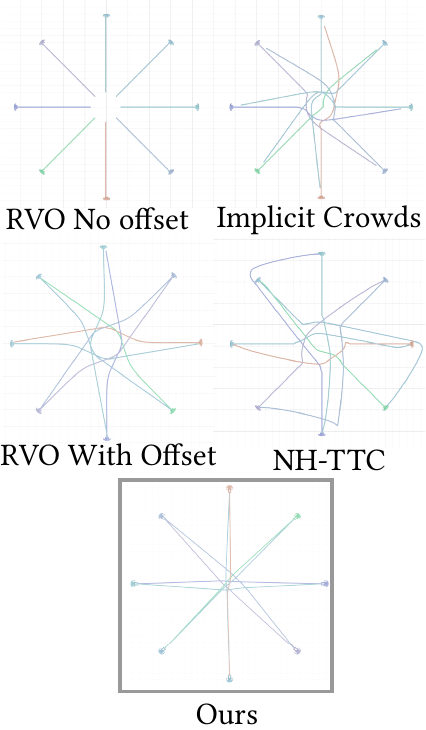}
  \caption{8 agents} 
  \label{fig:8personcomparisons}
  \end{subfigure}
  \caption{On the left, circles denote the start position of the agents and starts denote the desired end position. Only our method lets agents reach their desired ends without any locking. On the right, each agent wants to traverse to the directly opposite end of the circle. Our method leads to the most efficient trajectories.}
\end{figure}

Our method treats each agent as an individual space time curve with only three requirements: a starting position and time and an ending position (at indeterminate time). We use a globally supported, differentiable LogSumExp smooth distance in space-time to guarantee collision free trajectories and solve the resulting problem using an interior-point technique. For small to medium scale crowds, our method outperforms current state-of-the-art methods in simple scenarios (\autoref{fig:3personcomparisons} and \autoref{fig:8personcomparisons}) and more complicated, constrained environments. Our agents exhibit anticipatory behaviors such as slowing and waiting to let others pass and makes no assumptions about agents having identical mass or other physical properties. This enables planning of intricate scenes with environmental constraints such as the one described in our introduction. 
\vspace{1em}
\begin{table*}[ht]
    \begin{tabular}{l l l l l l l l l}
    \hline
    \rowcolor{greycolor}
    Features & Ours & (1) RVO & (2) IC & (3) NH-TTC & (4) RC & (5) CC & (6) STPL & (7) AABS\\
    \hline
    Asymmetric Interactions & \cellcolor{green!25}Y & \cellcolor{red!25}N & \cellcolor{black!25}P & \cellcolor{black!25}P & \cellcolor{red!25}N & \cellcolor{red!25}N & \cellcolor{red!25}N & \cellcolor{red!25}N\\
    \hline
    \rowcolor{greycolor}
    Extremely Constrained Environments & \cellcolor{green!25}Y & \cellcolor{red!25}N & \cellcolor{red!25}N & \cellcolor{red!25}N & \cellcolor{green!25}Y & \cellcolor{red!25}N & \cellcolor{black!25}P & \cellcolor{green!25}Y\\
    \hline
    Smooth Local Interactions & \cellcolor{green!25}Y & \cellcolor{red!25}N & \cellcolor{green!25}Y & \cellcolor{red!25}N & \cellcolor{green!25}Y & \cellcolor{red!25}N & \cellcolor{green!25}Y & \cellcolor{red!25}N\\
    \hline
    \rowcolor{greycolor}
    Intuitive Control Parameters & \cellcolor{green!25}Y & \cellcolor{green!25}Y & \cellcolor{red!25}N & \cellcolor{black!25}P & \cellcolor{red!25}N & \cellcolor{red!25}N & \cellcolor{red!25}N & \cellcolor{green!25}Y\\
    \hline
    Multiple Agents & \cellcolor{green!25}Y & \cellcolor{green!25}Y & \cellcolor{green!25}Y & \cellcolor{green!25}Y & \cellcolor{green!25}Y & \cellcolor{green!25}Y & \cellcolor{red!25}N & \cellcolor{green!25}Y\\
    \hline
    \rowcolor{greycolor}
    Code Available Online & \cellcolor{green!25}Y & \cellcolor{green!25}Y & \cellcolor{green!25}Y & \cellcolor{green!25}Y & \cellcolor{green!25}Y & \cellcolor{red!25}N & \cellcolor{red!25}N & \cellcolor{red!25}N\\
\end{tabular}
\caption{Compare (1) RVO (~\cite{van2011reciprocal}), (2) Implicit Crowds (\cite{karamouzas2017}), the brand new time-to-collision method (3), NH-TTC (\cite{nhttc}), (4) Repulsive Curves (\cite{repulsivecurves}), (5) Continuum Crowds  (\cite{treuille2006continuum}), (6) Space Time Planning With Parameterized Locomotion Control (\cite{SpacetimePlanning}), and (7) Modular Framework for Adaptive Agent Base Steering (~\cite{singh2011modular}). Y - feature is available, N - feature is not possible, P - feature might be possible, but not demonstrated.}
\label{table:crowds_vs_others}
\end{table*}

\section{Method}
At a high level we are influenced by the notion of correlated equilibrium. 
In a correlated equilibrium solution to a non-cooperative game, an "oracle" chooses a strategy for each player, and no player has any reason to deviate from the chosen strategy assuming others do not deviate either. 
The result is an equilibrium solution which maximizes collective utility.

Our method acts as the oracle of the scene and plans agent trajectories in a way that collectively maximizes the utility of the entire scene. Additionally, unlike previous approaches which `pre-set' trajectories for agents (often done manually), we allow our agents to find their own utility-maximizing trajectories. Lastly, real life agents have different sizes, masses, and personalities, which affect the agents' utilities, and therefore its path as well. Our local-global path planning approach allows for these nuances. 

\begin{figure}[h]
\includegraphics[width=\columnwidth]{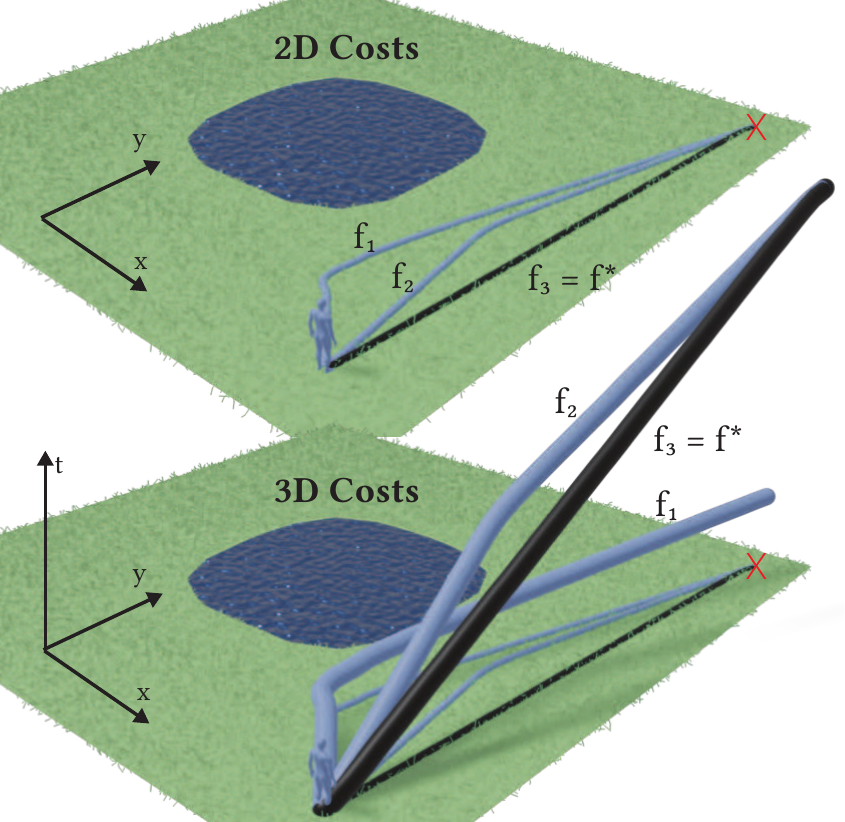}
\caption{ Top: A path through space can be represented as a curve embedded in $R^2$. Every path has a scalar cost (utility) value. For a cost function that minimizes distance travelled, $c_1$ and $c_2$ are inoptimal paths from the start to the goal (red X), but $c_3$ is optimal. Bottom: A path through space and time can be represented as a curve in $R^3$. A space-time rod (as shown here) is simply a 3D curve with a collision radius.} 
\label{fig:2D-3D-Paths}
\end{figure}

\subsection{Paths Map To Utility}\label{ssec:space-time-paths}
As shown in \autoref{fig:2D-3D-Paths}, for an agent, each path from the start location to the end location through space maps to some scalar utility value. For example, if the agent's utility minimized distance traveled, a straight line from start to end would prove to be the optimal path. In~\autoref{fig:2D-3D-Paths}, the 2D x-y plane describes the domain of spatial motion for agent paths. We make the additional observation that agents do not simply move through space, but also move through time which is denoted by the vertical z-axis. So, in 3D space-time, the agent's motion through space ($x,y$) and time ($t$) is described by a 3D curve which corresponds to a utility. For any given moment in time, the projection of the curve onto the x-y plane gives us the agent's location.

\begin{wrapfigure}{r}{110pt}
\centering
\vspace{-1\baselineskip}
\includegraphics[width=110pt]{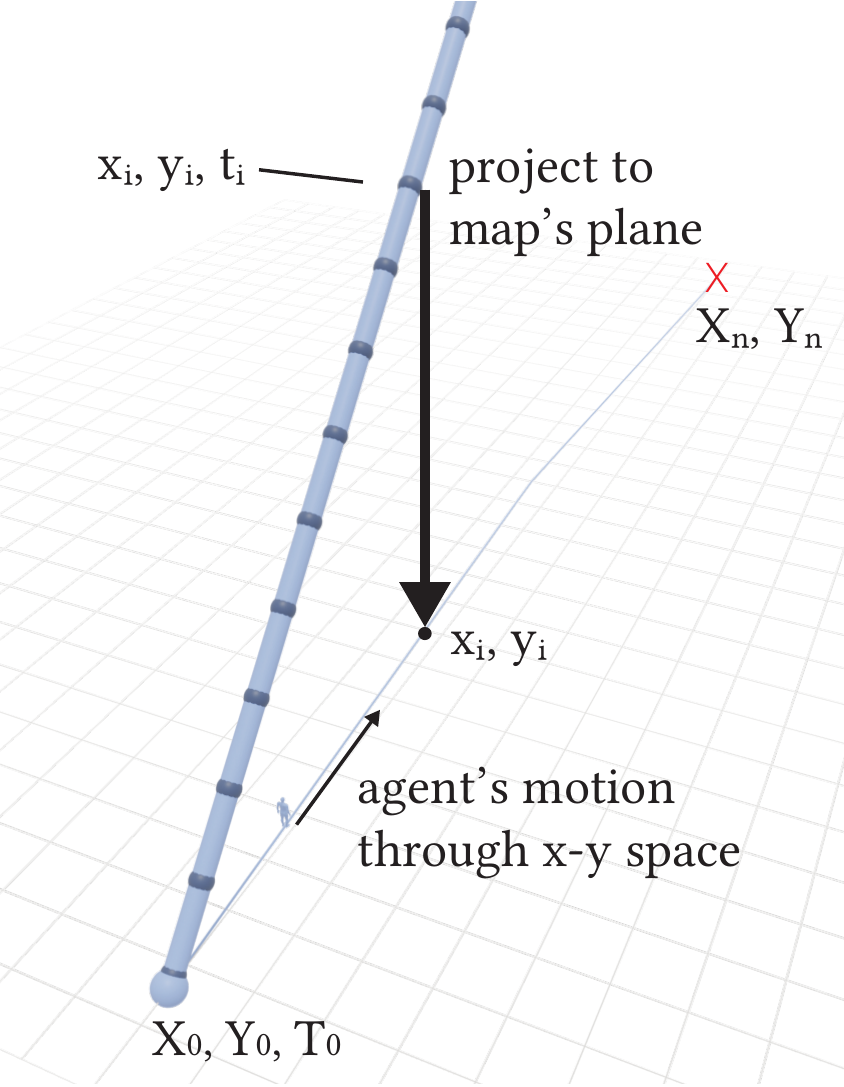}
\caption{This space-time rod describes a discrete space-time trajectory curve made up of many nodes. Our agent moves along the projection of the rod from its start to end, taking into account the boundary constraints of the problem.}
\label{fig:3D-render-didactic}
\vspace*{-1.4\baselineskip}
\end{wrapfigure}

Let us consider a simple case where there are no other agents in the scene and the terrain is flat and free of obstacles. The agent, indexed henceforth by subscript $a$, wishes to maximize its utility (by minimizing cost, $\bar\Psi$); the tilde symbol indicates the variable or function is continuous, not discrete. Lower case variables indicate a single agent. Upper case indicates the variable aggregates all agents. The motion of the agent through space and time denoted by the agent's 3D space-time curve $\conts_a$ embedded in $R^{x,y,t}$ requires constraints. Fortunately all our constraints are linear, so we lump them together into $\tilde{B}$ for now. Put together, the full optimization for a single agent $a$ is
\begin{subequations}
\begin{alignat}{2}
f_a^* &= \min \; \int_{\conts_a} &&\tilde{\psi}(\conts_a) \text{d}\conts_a \label{eq:contC} \\
& s.t. && \tilde{B}\conts_a \leq \mathbf{0}.
\end{alignat}
\end{subequations}
The cost,~\autoref{eq:contC}, for agent $a$ is integrated over the trajectory of the agent $(\conts_a)$. The specific nature of the cost function determines the agents' behavior based on its characteristics. An agent might have a preferred walking speed, or a stubbornness factor, or a radius of comfort, all of which (and more) can be encoded into components of $\tilde{\psi}$. In order to solve this optimization, we must first discretize the agent's trajectory into the discrete curve $\diss_a$ shown in~\autoref{fig:3D-render-didactic}. We also discretize the cost $\tilde{\psi}$ into separate intra-agent costs, which solely affect the path of one individual, and interaction costs, which can affect multiple agents.

\subsection{Discretizing}
Our agent's cost function takes in the agent's \textbf{discrete} 3D space-time curve $\diss_a$ as input and outputs a scalar cost for that path, $f_a$. We descretize the agent's continuous $\conts$ into a piecewise linear curve described by $n+1$ nodes $\diss_a =[(x_a^0, y_a^0, t_a^0),\allowbreak...,\allowbreak(x_a^i, y_a^i, t_a^i),\allowbreak....,\allowbreak(x_a^n, y_a^n, t_a^n)]$ for nodes $i=0..n$ connected sequentially by edges. We must also discretize our constraints. First, even though time is a variable in our formulation,~\autoref{eq: forwardsInTimeConstraint} ensures the agent cannot move backwards in time. Second, the agent has a start location $(x_a^0, y_a^0)$ and start time $t_a^0$ denoted in~\autoref{eq: startConstraint}.  Third,~\autoref{eq: endConstraint} sets a goal or an end location $(x_a^n, y_a^n)$. Lastly, the agent cannot take an infinite amount of time, so~\autoref{eq: maxTimeConstraint} bounds the agent by a max time $T_a^{max}$. Putting all of this together, we can re-write the optimization for agent $a$ with the discrete generalized cost function $\psi$ as
\begin{subequations}
\begin{alignat}{2}
f_a^* &= \min \; \sum_{i=0}^{n} && \psi(\diss_a) \label{eq:dissC} \\
 & s.t. && t_a^i \leq t_a^{i+1} \label{eq: forwardsInTimeConstraint} \\
& && (x_a^0, y_a^0, t_a^0)= \mathbf{(x_a, y_a, t_a)^{start}} \label{eq: startConstraint}\\
& && (x_a^n, y_a^n) = \mathbf{(x_a, y_a)^{end}} \label{eq: endConstraint}\\
& && t_a^n \leq T_a^{max} \label{eq: maxTimeConstraint}.
\end{alignat}
\end{subequations}
Now with a template for our optimization problem, we can replace the generalized cost function with specific discrete costs. 
Our specific cost functions are derived from observation of real behavior. These behavioral observations are divided into two categories, intra-agent costs, and interactions costs. Intra-agent cost functions only look at one agent's path at a time. Interaction cost functions include avoiding collisions with other agents, collisions with static obstacles such as walls or furniture, grouping behavior within friends, asymmetric behavior based on the mass, size of the agents, or other characteristics.

\subsection{Intra-Agent Costs}
Since these costs apply to a single agent, we calculate intra-agent costs for arbitrary agent $a$ and later we show how to sum the costs over the entire scene. The path of our agent, $\diss_a$, is comprised of of $n+1$ nodes indexed by $i=1..n$. Each node $(x^i(\diss_a),y^i(\diss_a),t^i(\diss_a))$ is comprised of the agent's spatial $(x,y)$ coordinates and time $(t)$ coordinates. For all cost functions, agent $a$ also has constant weighting terms, $K_a$, as well as constant characteristics such as mass, $m_a$, and preferred end time, $T_a^p$.

\subsubsection{Intra-Agent Kinetic Cost}
Our agent $a$ will try to minimize energy expenditure. Given an agent's space-time curve $\diss_a$, the kinetic energy cost of the agent over the curve can be written as
\begin{align}
&C_a^{K}(\diss_a, K_a^K) = K_a^K  \sum_{i=0}^{n} \frac{1}{2}m_a\frac{(\Delta \mathbf{x}_a^i)^T (\Delta \mathbf{x}_a^i)}{(\Delta t_a^i)^2}(\Delta t_a^i) \label{eq:kineticunsimplified}\\
&= K_a^K  \sum_{i=0}^{n} \frac{1}{2}m_a\frac{(x^{i+1}(\diss_a) - x^i(\diss_a))^2 + (y^{i+1}(\diss_a) - y^i(\diss_a))^2}{t^{i+1}(\diss_a) - t^i(\diss_a)} \label{eq:kineticsimplified}
\end{align}
where~\autoref{eq:kineticunsimplified} is the kinetic energy $\frac{1}{2}mv^2$ for curve segment $e_a^i = [x_a^i,y_a^i,t_a^i,x_a^{i+1},y_a^{i+1},t_a^{i+1}]$ integrated over total time travelled $\Delta t_a^i = t_a^{i+1} - t^i$. This simplifies into~\autoref{eq:kineticsimplified} where $K_a^K $ is the agent-wise weighting coefficient, $m_a$ is the mass of the agent and $\diss_a$ is the discretized path curve (our input variable) made up of $n+1$ nodes.

\subsubsection{Intra-Agent Acceleration Cost}
In order to penalize acceleration in agent $a$'s trajectory, the acceleration cost 
\begin{align}
C_a^A(\diss_a, K_a^A) \;= K^A_a \sum_{i=1}^{n-1} \frac{1}{2}(\theta^i_a)^2 \label{eq:accCost}
\end{align}
measures the curvature of $\diss_a$ through the discretized acceleration cost where angle $\theta$ is the angle between two piece-wise linear segments of the curve computed using the stable arctan function $arctan2$. The angle is $\theta_a^i \; =\allowbreak arctan2(\frac{\|(\diss_a^{i+1} - \diss_a^i) \times (\diss_a^i - \diss_a^{i-1}) \|}{(\diss_a^{i+1} - \diss_a^i)^T(\diss_a^i - \diss_a^{i-1})})$ where $\diss_a^i = [x^i, y^i, t^i]$ for each node in the curve. 

\subsubsection{Intra-Agent Preferred End Time Cost}
Any agent $a$ has a preferred end time, $T_a^p$ at which they expect to reach the end position. Sometimes this is the same as the max end time $T_a^{max}$ by when the agent is \textbf{required} to be at the end positions, but sometimes the preferred end time $T^p_a$ can be sooner. Deviation from the preferred end time is modeled as a quadratic cost
\begin{align}
C_a^T(\diss_a, T_a^p, K_a^T) \; = K_a^T \frac{1}{2} (t_a^n -  T_a^p)^2 \label{eq:ptCost}
\end{align}
incentivizing the agents to arrive at their preferred end time $T_a^p$ by keeping the actual end time for each agent, $t_a^n$, close to $T^p_a$. 

\subsubsection {Regularizing Cost}
We find that adding a regularizing term to penalize extremely short time segments improves the overall quality of the paths by reducing near instantaneous motions in time. To penalize very fast agent motion, we use the regularizing term
\begin{align}
C_a^R(\diss_a, K_a^R) \; = K_a^R \sum_{i=0}^{n}  \left(\frac{\frac{t^n_a}{n}}{t^{i+1}_a - t^i_a} \right). \label{eq:regCost}
\end{align}
This is important because it allows us to feed the solver an initial space-time curve such as the ones in~\autoref{fig:initial-spacetime-curve} with many near instantaneous agent motions, and the solver is able to optimize the final space-time curve to a much more reasonable trajectory.

\subsection{Interaction Cost}
\label{ssec:interactionCost}
\begin{figure}[h]
\includegraphics[width=\columnwidth]{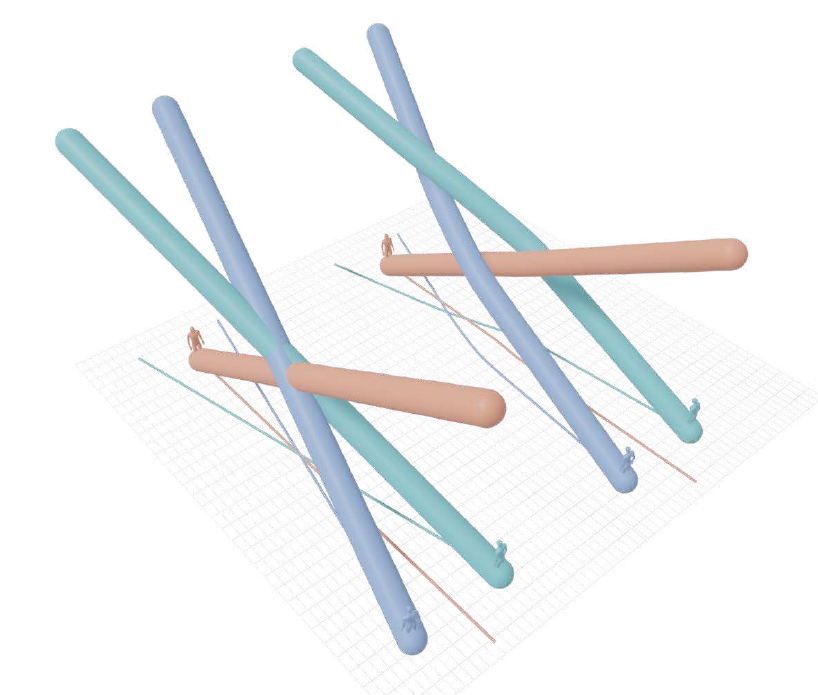}
\caption{ Resolving intersections between agents space-time rods resolves collisions in the scene since no agent shares the same $x,y,t$ location at any point in the scene.} 
\label{fig:3-person-setup}
\end{figure}

Interactions include agent-environment interactions and agent-agent interactions for which we must introduce a new arbitrary agent indexed by $b$ where $a \neq b$. All our interactions depend on each agent's `radius of comfort' (or collision radius), denoted by $r_a, r_b$ for agents $a$ and $b$. The collision radius might be based on size, or a combination of size and personality: for example, people stay away from angry people. We encode these agent characteristics into our cost functions by extruding circle with radius $r_a$ along the agent's path $\diss_a$ thus forming a rod. In a scene with multiple agents, as long as no two rods are intersecting, the scene is collision free as shown in~\autoref{fig:3-person-setup}. The collision radius ensures that no two agents are in the same place at the same time. So even though the forces between the two space-time rods might be symmetric, the agents' response to these forces (change in path) leads to asymmetric interactions. For static object interactions, the environment boundary is encoded into boundary vertices $b_v$ and boundary elements $b_e$. Let us examine how we detect which rods are intersecting and how we deal with our three interaction types: agent-agent collisions, agent-agent groupings (friendships), and agent-environment collisions.

\subsubsection {Collision Interaction Cost}
\begin{wrapfigure}{r}{0.33\columnwidth}
\includegraphics[width=0.33\columnwidth]{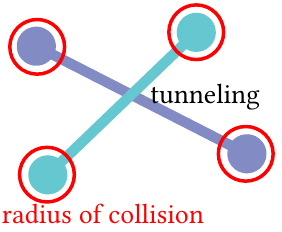}
\caption{Tunneling artifact.}
\label{fig:tunneling}
\end{wrapfigure}

\begin{figure}[h]
\includegraphics[width=\columnwidth]{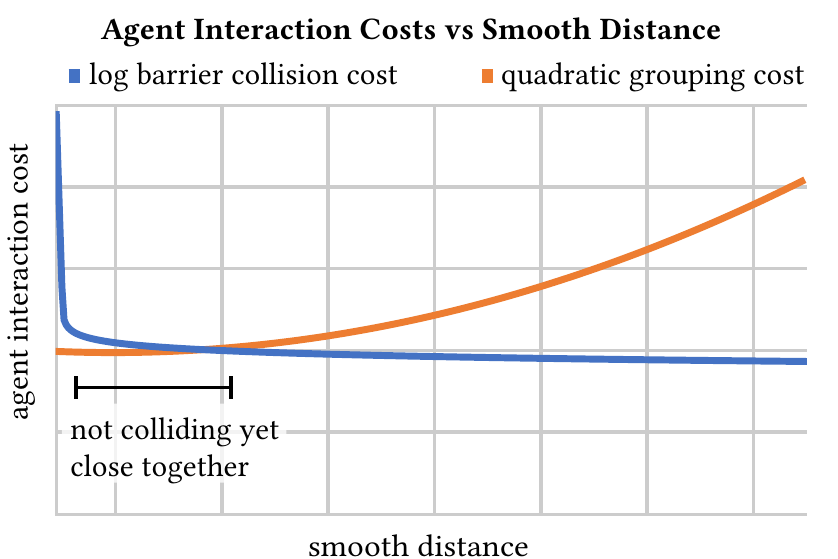}
\caption{ The log barrier energy quickly tends towards infinity as the minimum distance approaches the collision radius.} 
\label{fig:log-barried-function}
\end{figure}

Given a pair of agents $a,b$ with 3D space-time curves $\diss_a, \diss_b$ and collision radii $r_a, r_b$, we densely sample each space-time curve uniformly. Next, we calculate the minimum smooth distances between the upsampled centerlines using the method described in~\autoref{sssec:smooth-distance}. As long as the smooth minimum distance $\text{d}$ between the upsampled rods $u_a = \text{upsample}(\diss_a), u_b=\text{upsample}(\diss_b)$ is further apart than $r_a + r_b$, collisions will not occur. We implement this non-linear constraint on the agent paths as a log barrier energy:
\begin{align}
& C_{a,b}^C(\text{d}(\alpha, \text{upsample}(\diss_a), \text{upsample}(\diss_b)), r_a, r_b, K_{a,b}^C) \nonumber \\
& = -K_{a,b}^C \log(-(r_a + r_b) + \text{d})). \label{eq:collisionCost}
\end{align}
where $K_{a,b}^C$ is the pair-wise weighting coefficient on the energy. The intuition behind a log barrier energy is explained in~\autoref{fig:log-barried-function}. 
Upsampling the trajectories prevents tunneling artifacts (\autoref{fig:tunneling}) between edges during collision detection. The collision cost increases exponentially as the minimum distance between the two agent rods approaches the sum of the collision radii, so collisions are exponentially penalized. By summing this cost with the kinetic energy term, which incentivizes agents to move at a constant velocity, we get smooth local collision resolution as shown in~\autoref{fig:3-person-setup}.

\vm{We employ two methods for sparsifying interactions. First, we use a 3D tree structure to store agent trajectory nodes (analogous to a Bounded Volume Hierarchy BVH) for broad-phase collision detection between trajectories. The tree-based broad phase reduces collision detection costs from $O(n^2)$ to $O(nlogn)$. Second, if two agents are known to be far apart, we entirely avoid the collision detection step between those two agents thus reducing costs to $O(n)$ in the best case. Using a combination of tree based broad phase along with manual denotation of interacting agents, we find that our method scales nearly linearly $O(n)$as the number of agents in the scene increases.}

\subsubsection {Grouping Interaction Cost}
While the collision resolution term pushes agent paths apart, we introduce a grouping term which pulls paths together
\begin{align}
& C_{a,b}^G(\text{d}(\alpha, \text{upsample}(\diss_a), \text{upsample}(\diss_b)),r_a, r_b, K_{a,b}^G) \nonumber \\
& = K_{a,b}^G(\text{d} - (r_a + r_b))^2. \label{eq:groupingCost}
\end{align}
This quadratic grouping term allows us to model scenarios in which friends going in the same direction tend to stick together as shown in~\autoref{fig:3PersonAsymmetrics}, or scenarios in which an agent needs to deliver a message to another agent who is out of the way from his final destination such as~\autoref{fig:pondScenes}. The weighting term $K_{a,b}^G$ controls the desire of the agents $a,b$ to group together.

\subsubsection{Agent-Environment Interaction}
We surpass previous methods in three ways in terms of agent-environment interactions. Firstly, our method works on agents traversing highly constrained environments such as~\autoref{fig:roomba-maze},~\autoref{fig:subway-platform}, or~\autoref{fig:cornMaze}. Secondly, our method allows agents to explore routes when multiple paths are available and choose the most optimal path using a novel modification to Djikstra's algorithm. Thirdly, our path planning step is topology aware, so this method will work on more fascinating terrain.

Environment maps are stored as 2D mesh files with vertices and faces. We pre-compute the vertices and edges around the boundaries of the map and any static obstacles into $b_v$ and $b_e$ as a one-time preprocessing step. Next, we compute the minimum smooth distance between the map boundary edges and the agent rod. Luckily since static obstacles are fixed in space for all time, we can ignore the third dimension and only compute the 2D minimum distance, $\text{d}(\alpha, \diss_a, b_v)$ from the rod's projection onto the map and the map's boundary edges. Next we pass the smooth distance $\text{d}$ into the log barrier function
\begin{align}
C_a^M(\text{d}(\alpha, \text{upsample}(\diss_a), b_v), r_a)  = -K^M_a log(-r_a + d) \label{eq:mapCost}
\end{align}
where $K^M_a$ is the energy coefficient. Like in the agent-agent collision function (\autoref{eq:collisionCost}), we can use the collision radius $r_a$ for each agent since the initial agent paths are sufficiently far enough away from static obstacles such that $d>r_a$. The coarseness of the map mesh does not impact the smoothness of agent path; however, Djikstra's algorithm requires two vertices on the map which correspond to the agent's start and end positions to generate the initial path. Therefore, we set the nearest vertices to the given start and end points as the boundary conditions for the solve.

\subsection{Smooth Min Distance Implementation}
\label{sssec:smooth-distance}
The absolute minimum distance between agent paths is not a differentiable function. Therefore, we use a differentiable LogSumExp smooth minimum distance function to smoothly approximate the distance between two upsampled space-time curve vertices $u_a$ and $u_b$. See the supplementary material for derivatives of the function. The smooth minimum distance function is
\begin{align}
d(\alpha, u_a, u_b) = \frac{-1}{\alpha} \log\left( \sum_{i=0}^n \sum_{j=0}^n e^{-\alpha \|\diss_a^i - \diss_b^j\|} \right)
\end{align}
where the $\alpha$ term controls the numerical sensitivity of the distance which changes depending on the size of the environment mesh and the distances between the agents' path curves. A higher $\alpha$ leads to a more accurate minimum distance, but reduces numerical stability. Since this smooth minimum distance function is an underestimation of the true minimum distance, it is possible to get negative distances for very close objects.

We use \autoref{alg:alpha-heuristic} to update $\alpha$ during the solve to guarantee a usable (real, non-negative) smooth minimum distance. Our initial $\alpha_0$ for the scene is as low as possible, but large enough to guarantee that both our log barrier interaction costs are real and defined. If the initial $\alpha_0$ is too small, the distance underestimation is too great, and the log barrier costs are either complex or undefined, then the interior point solver will throw an error. We must choose a large enough $\alpha_0$ that the heuristic will provide an $\alpha$ that guarantees a real smooth distance values where $D > r_a + r_b$ to ensure real log barrier costs. For subsequent iterations, any real, non-negative smooth minimum distance is fine. 
\vspace{1em}
\begin{algorithm}
\caption{A heuristic to find a usable $\alpha$ to be used within the agent-collision and map interaction cost, gradient and hessian functions.}
\label{alg:alpha-heuristic}
\begin{algorithmic}

\Function{ALPHAHEURISTIC}{$\alpha_0,\diss_a, \diss_b$}
\State $\alpha \gets \alpha_0$
\State $D \gets d(\alpha, \diss_a, \diss_b)$
\While{$D \leq 0$}
  \State $\alpha \gets \alpha + 0.1*\alpha_0$
  \State $D \gets d(\alpha, \diss_a, \diss_b)$
\EndWhile
\State \Return $\alpha$
\EndFunction
\end{algorithmic}
\end{algorithm}

\subsection{Initial Paths}
\label{sssec:space-time-djistras}
Using an interior point solver requires that we find feasible initial trajectories in which the agents do not collide with other agents or any obstacles. \vm{The supplementary material shows our failed experiments in generating initial paths based on shortest distances. Some global path planner is necessary to generate feasible initial paths.} Initially intersecting space-time rods causes NaNs in the log barrier interaction costs which makes the interior point method intractable. \vm{In order to generate initial plausible paths to feed into the optimization, we sequentially run a modified Djikstra's algorithm for all agents traversing the environment mesh inspired by~\cite{treuille2006continuum}.} 

Djikstra finds the distance-weighted least costly path along the edges of the map mesh from the start location to the end location for a given agent. We avoid obstacles (environment boundary) by giving these edges, $b_e$, a prohibitively high traversal cost. We only consider map edges that have a distance from the map boundary at least greater than the agent's collision radius.
\begin{wrapfigure}{r}{0.6\columnwidth}
\centering
\vspace{-0.1\baselineskip}
\includegraphics[width=0.6\columnwidth]{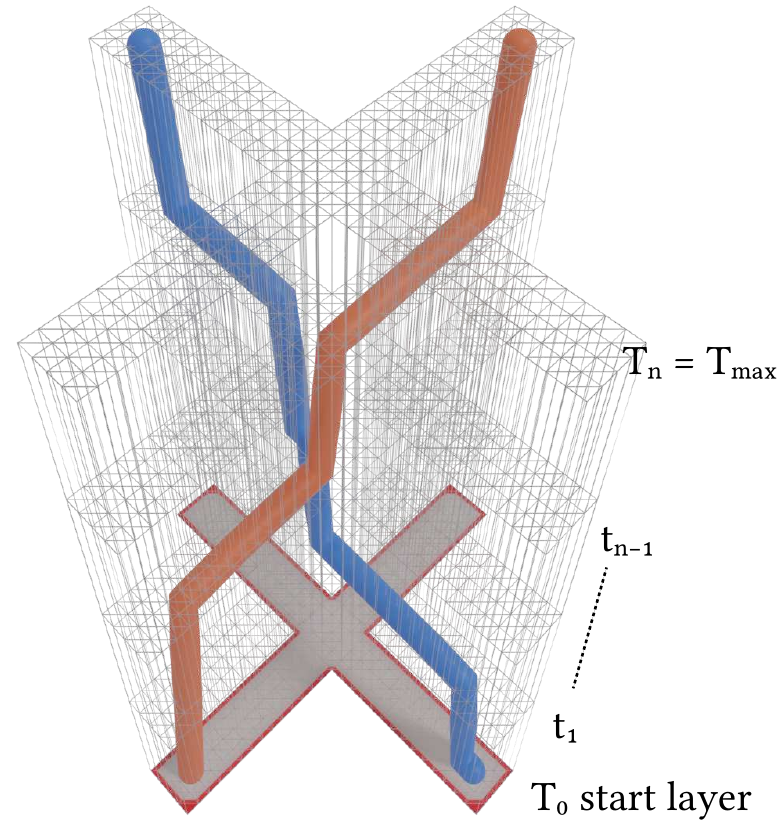}
\caption{ Running Djikstra's algorithm to set initial agent paths on the map. Red map edges indicate a higher traversal cost. Since we want to set a feasible initial path for each agent, we keep them away from the map edges as well as away from other agent's initial paths.}
\label{fig:djikstras-map}
\vspace*{-0.5\baselineskip}
\end{wrapfigure}

Furthermore, we ensure that every agent has a collision-free initial path by making it impossible for an agent's initial path to pass too close to the initial path of another agent. For each agent, upon finding an initial path, we wipe out all edges connected to the path within the agent's collision radius. For very constrained scenes, it becomes impossible to ensure collision free initial trajectories since agents often run out of traversible edges from $[x^0, y^0]$ to $[x^n, y^n]$. Since agents move through both space and time, we extrude our environment map into the $t$ dimension and create a pre-set number of layers from $t^0$ to $T^{max}$ as shown in~\autoref{fig:djikstras-map}. Solving this 3D space-time Djikstra's algorithm makes it a lot more feasible to find collision-free inital paths for all agents. Sometimes, the coarseness of the environment mesh, or the low number of layers in the 3D space-time map makes it impossible for Djikstra's algorithm to find a feasible initial path. In this case, either the environment map would need to be subdivided, more layers would need to be added, or the scene itself might be infeasible given the $T^{max}$ time constraints and the number and size of agents involved.
\begin{figure}[h]
\includegraphics[width=\columnwidth]{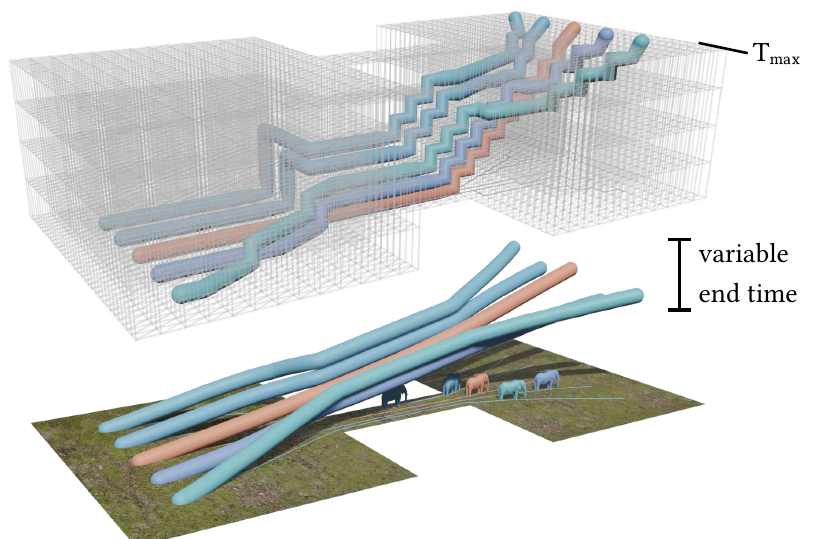}
\caption{ Top is the initial space-time curve output from Djikstra's path finding algorithm. Bottom is the final result of the solver with optimal paths. Agents are allowed to find their own optimal end-times while avoiding collisions in tightly constrained bottlenecks such as this.} 
\label{fig:initial-spacetime-curve}
\end{figure}
Additionally, although the space-time Djikstra's algorithm will provide feasible initial paths, they will have many kinks and sharp turns (\autoref{fig:initial-spacetime-curve}) indicating unnatural motion and thus cannot be used in the simulation directly. These get smoothed out during the optimization process resulting in much more realistic motion. \vm{It is possible that this proposed method to generate initial paths will affect the state of the final trajectories. For example, if an initial path passes through the left of an obstacle, it might be stuck in a local minimum even though an optimal trajectory might be through the right side. This drawback is somewhat mitigated by using Djikstra's search to pre-compute the least costly paths. Either way, whether or not the solver produces a globally minimum outcome, the trajectories are guaranteed to be smooth, viable and collision free.}

\subsubsection{Optimization}
Now aggregating over all the agents in the scene, $\|A\|$, we put intra-agent cost functions Equations 4-7
together with interaction costs Equation 8-10
and construct an optimization problem for our scene. We aggregate boundary conditions for each agent and denote them with capital letters to show that they apply to the entire scene. The optimization problem
\begin{subequations}
\begin{alignat}{2}
&f^* = \min \;&& \sum_{a=1}^{\|A\|} C_a^K(\cdot) + C_a^A(\cdot)+ C_a^T(\cdot) + C_a^R(\cdot) + C_a^M(\cdot)+ \\
& && \sum_{\substack{a,b \epsilon \\ \{1..\|A\|\} \\ | a \neq b }} C_{a,b}^G(\cdot) 
+ C_{a,b}^C(\cdot) \\ 
& s.t. && T^i \leq T^{i+1} \\
& && (X^0, Y^0, T^0)= (X, Y, T)^{start} \\
& && (X^n, Y^n) = (X, Y)^{end} \\
& && T^n \leq T^{max}
\end{alignat}
\end{subequations}
is solved using a quasi-newton interior point method. An outer loop over this optimization is required for a proper log-barrier method as explained by~\cite{NoceWrig06}. The full pseudocode overview of our method's outer loop,~\autoref{alg:fullmethhod}, illustrates this. The supplementary material for our method include the subroutines, the calculations for the gradients and hessian approximations of our costs, intuition for controlling agent behavior through function weights. 
\vspace{1em}
\begin{algorithm}
\caption{A full pseudocode overview of our method with subroutines provided in the supplementary material.}
\label{alg:fullmethhod}
\begin{algorithmic}
\Variables
 \State $OuterIts\geq1$, outer loop iterations
 \State $\mu, c = 0.75$, log-barrier coeff and its decrement factor
 \State $\alpha_0, cutoff=0.2$, initial alpha, Hess sparsifying cutoff
 \State $T^{max}$, max time constraint for agents
 \State $b_e, b_v$, environment edges and vertices.
 \State $K$, cost function weights for agents
 \State $R$, collision radius for agents
 \State $T$, preferred end times
 \State $M$, agent masses
\EndVariables
\\
\IPVariables
 \State $MaxIts$, max iterations
 \State $B$, trajectory boundary conditions 
 \State Stop if $StepSize>10^{-2} (meters)$, norm of solver step 
 \State Stop if $FirstOrderOpt>10^{-2} (meters)$, first order optimality criteria 
\EndIPVariables
\\
\Require @COSTS, @GRADS, @HESS (supplementary material)
\\
\Function{FINDOPTIMALPATHS}{}
  \State $[\disS, B] \gets DJIKSTRASPREPROCESS(R, T, b_v)$
  \State $\disS \gets$ increment $t$ by $\epsilon$ to ensure $t_{i+1} > t_i$
  \While{OuterIts>0}
    \State $\disS \gets IPSOLVER(\disS, B, @COSTS, @GRADS, @HESS)$
    \State $\mu \gets c\mu$
    \State $OuterIts \gets OuterIts - 1$
  \EndWhile
  \State \Return $\disS$
\EndFunction
\end{algorithmic}
\end{algorithm}

\begin{figure*}[ht]
\includegraphics[width=\textwidth]{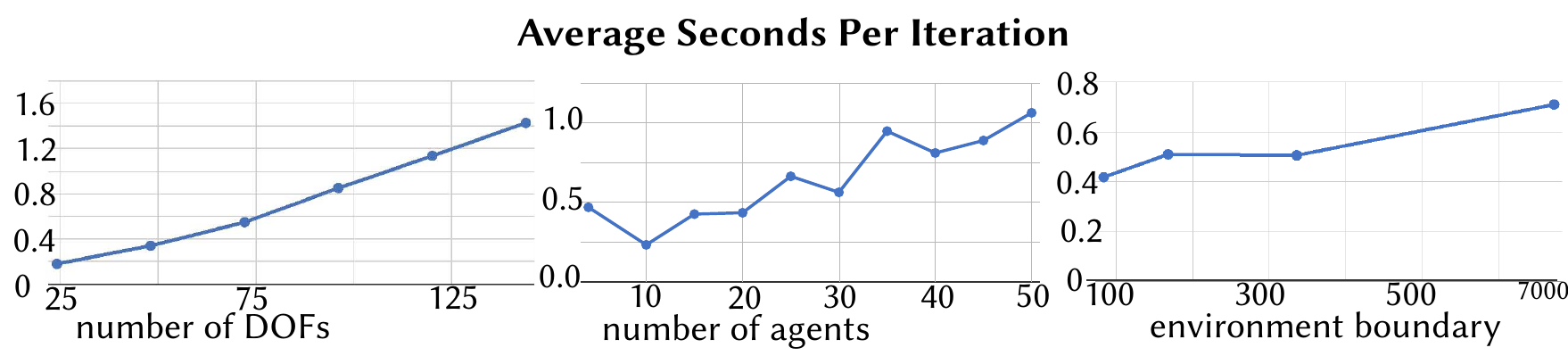} 
     \caption{(left) we varying number of agents with 300 DOFs each. (middle) we vary the number of DOFs/agent from 90 to 540 on a fixed scene with 8 agents. (right) we vary DOFs in the environment boundary from 2520 to 20160 on the corn maze mesh with 3 agents of 600 DOFs each in the scene. \vm{Scaling tests show near linear performance in the worst-case and linear performance in the best case.}}
     \label{fig:timings}
\end{figure*}

\section{Results}
\begin{figure*}[ht]
\includegraphics[width=\textwidth]{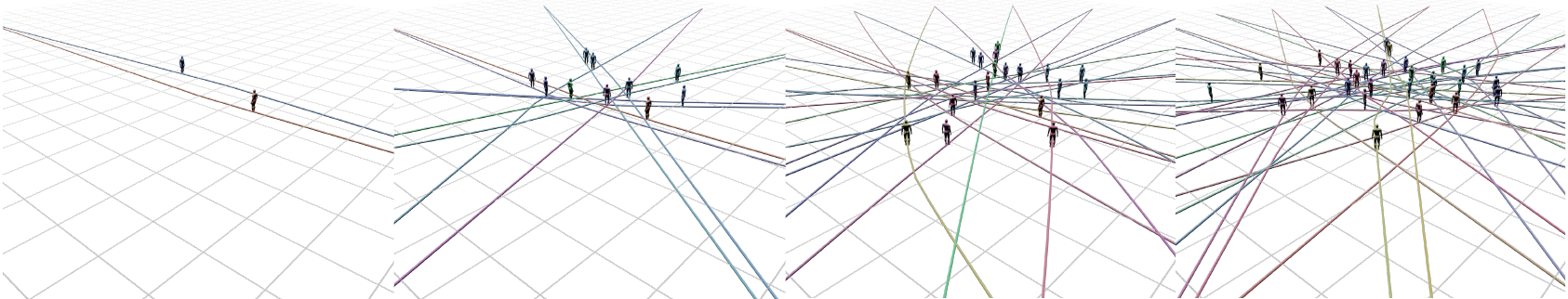}
\caption{Circles of agents.} 
\label{fig:circle}
\end{figure*}

\begin{figure*}[ht]
\includegraphics[width=\textwidth]{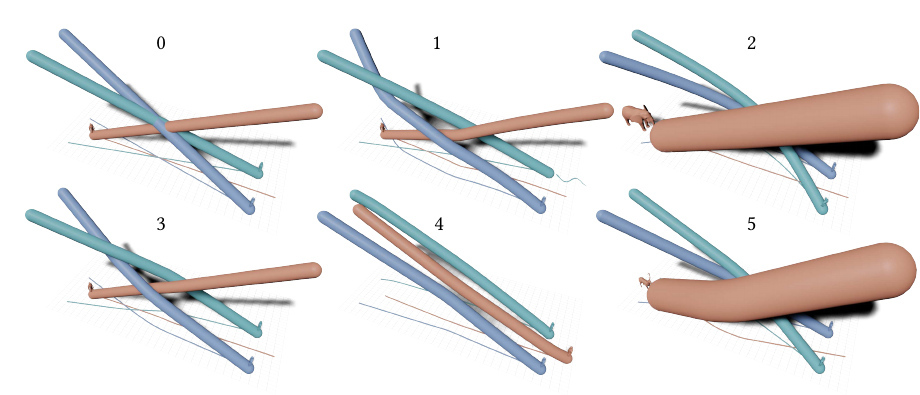}
\caption{(0) No interactions. (1) Mass (analogous to stubbornness) weighted asymmetric interactions. (2) Size and mass weighted interactions. (3) Symmetric interactions.(4) Grouping friends together.  (5) Size based interactions.} 
\label{fig:3PersonAsymmetrics}
\end{figure*}

\vm{THe performance of our method relies on (1) the number of agents in the scenes and (2) the complexity of the agent's paths through space and time. We push along each of these axes separately in this section. We show extremely complex maze-like environments with bottlenecks, as well as large scenes with tens (to hundreds) of agents.} In addition to the comparisons shown in the related works, we include the obligatory circles of agents~\autoref{fig:circle}. Notice that our agents take smooth, natural trajectories rather than the strange spinning motions demonstrated by other methods. Next, in~\autoref{fig:3PersonAsymmetrics} we highlight the several different types of asymmetric interactions supported by our method. We show the naive case where agents collide. We show standard symmetric collision between agents. We show a stubborn snake which forces the scared humans to move out of its path. We show a large elk who is still scared of the humans so it changes its trajectory just as the human agents change theirs. We show a large and stubborn elephant that goes straight to its destination while the humans have to significantly alter trajectories to avoid it. Finally, we show a scenario where two friends stick closer to each other on their way to their final destination. 

\begin{figure}[h]
\centering
  \begin{subfigure}[t]{0.49\columnwidth}
  \includegraphics[width=\columnwidth]{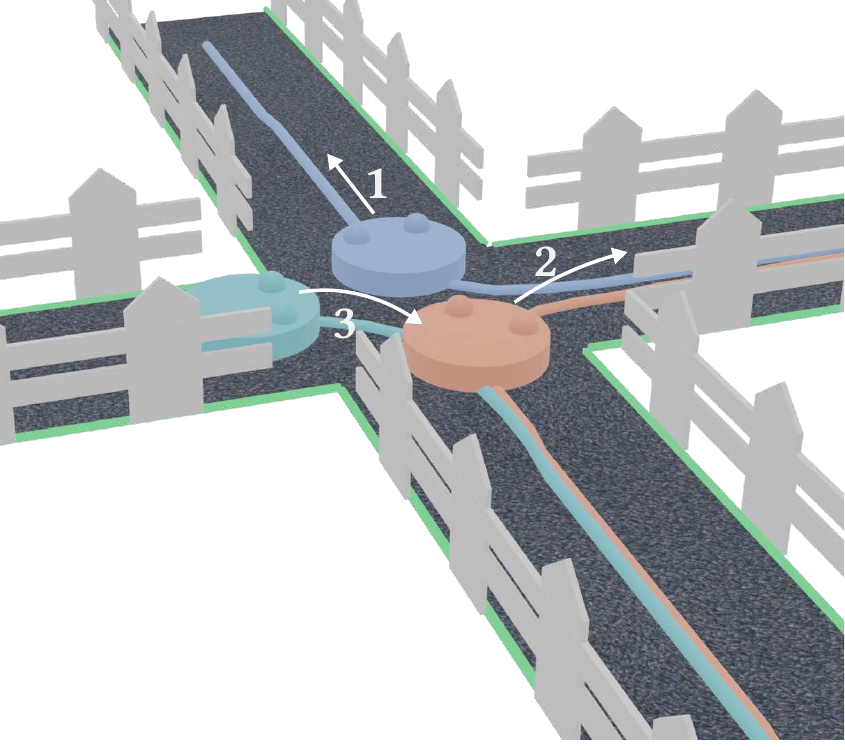}
  \caption{Maze of roombas.} 
  \label{fig:roomba-maze}
  \end{subfigure}
  \begin{subfigure}[t]{0.49\columnwidth}
  \includegraphics[width=\columnwidth]{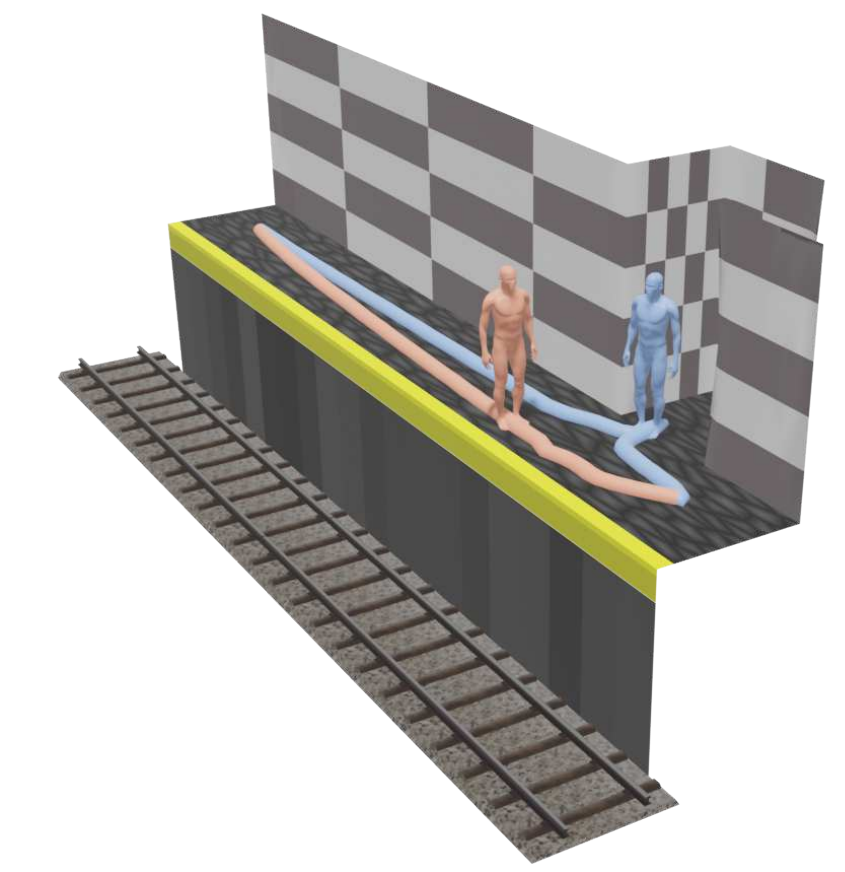}
  \caption{Subway platform} 
  \label{fig:subway-platform}
  \end{subfigure}
  \caption{In both (a) and (b) only one agent can move through the intersection at a time.} 
  \label{fig:highlyconstrained}
\end{figure}

\begin{figure}[h]
\includegraphics[width=\columnwidth]{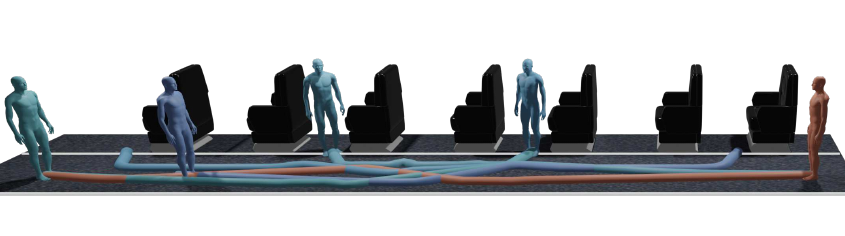}
\caption{Chaos in first class. All the passengers are in incorrect seats.} 
\label{fig:airplane}
\end{figure}

In~\autoref{fig:highlyconstrained} and~\autoref{fig:airplane}, we show our method works in extremely tight environments. RVO and Implicit Crowds show that they can navigate environmental constraints with a lot of manual pre-processing, but nowhere near as tight as our examples. Meanwhile, NHTTC does not implement environmental constraints altogether. We show that our method works in extremely constrained scenarios such as a subway tunnel or a tight warehouse of robots where each lane is big enough to accommodate only a single agent and agents have to take turns to pass through to avoid locking.

Although our method is built to handle asymmetric interactions and agents in highly constrained spaces, we also show several other features of our method. First, our agents have flexible arrival times. In~\autoref{fig:initial-spacetime-curve} we show a bottleneck where all the agents simply cannot arrive at their destinations by their preferred end times.  This feature serves as a counterpoint to the idea of using fixed start and end boundary conditions to model trajectory curves. In~\autoref{fig:airplane} we show a highly constrained bottleneck of agents trying to get to their seats on an airplane.

\begin{figure}[h]
\includegraphics[width=\columnwidth]{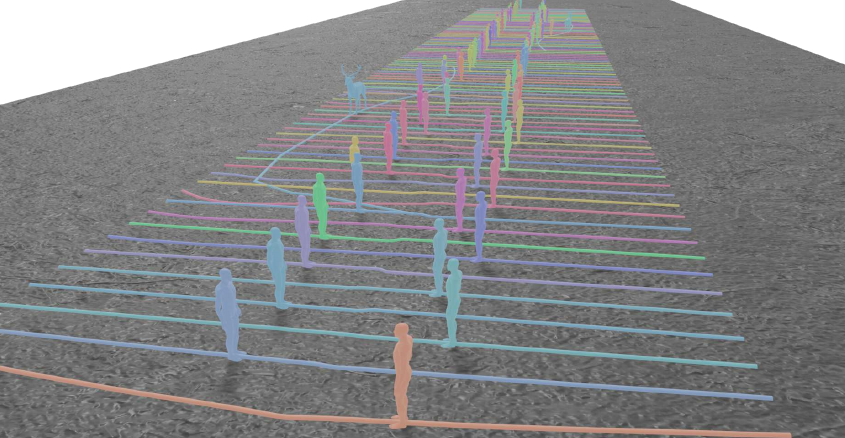}
\caption{A large scene with 102 agents moving to opposite sides of a meadow while a couple of deer avoid them.} 
\label{fig:102Agents}
\end{figure}

\begin{figure}[h]
\centering
  \begin{subfigure}[t]{0.45\columnwidth}
  \includegraphics[width=\columnwidth]{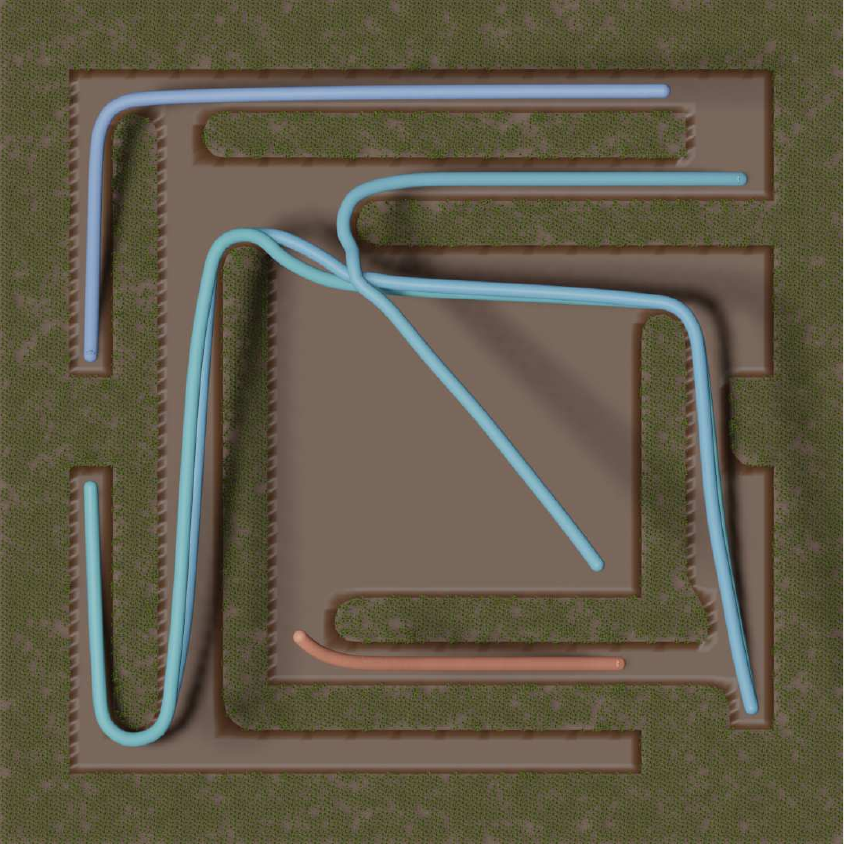}
  \caption{Corn maze}
  \label{fig:cornMaze}
  \end{subfigure}
  \begin{subfigure}[t]{0.45\columnwidth}
  \includegraphics[width=\columnwidth]{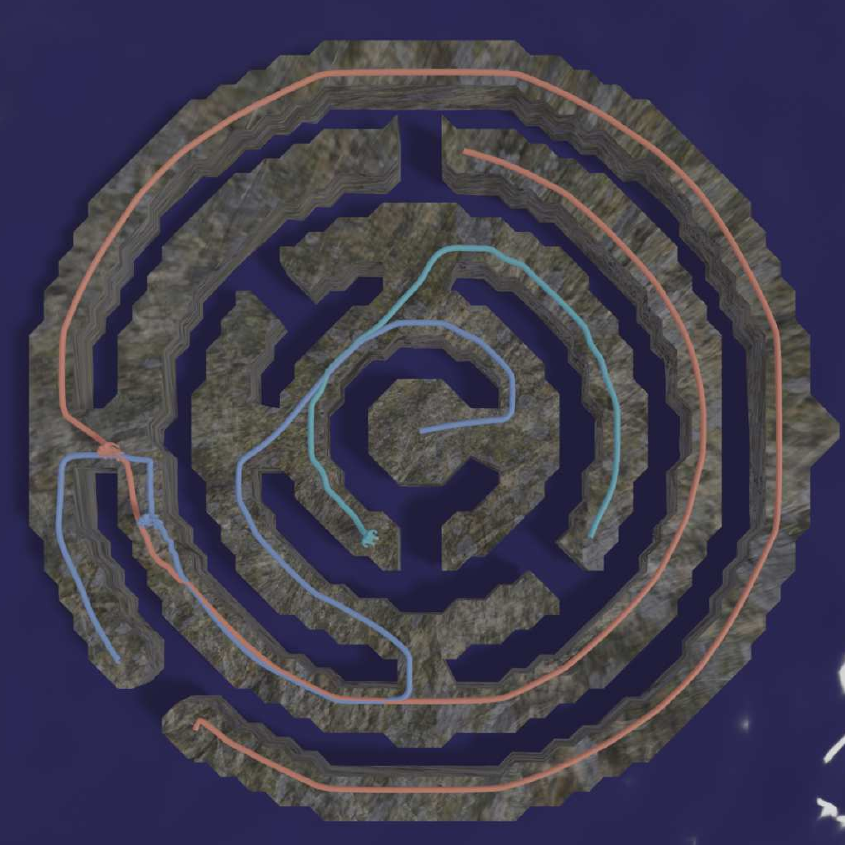}
  \caption{Circle maze}
  \label{fig:circleMaze}
  \end{subfigure}
  \caption{(a) large cornmaze with eight agents all trying to get to different locations through the maze. (b) a circular shaped maze where agents need to navigate while avoiding collisions.} 
\end{figure}

Additionally, we show the navigability of agents in larger constrained environments in~\autoref{fig:cornMaze} and~\autoref{fig:circleMaze}. Agents are able to navigate the mazes while avoiding collisions with each other at the bottlenecks.

\begin{figure}[h]
\includegraphics[width=\columnwidth]{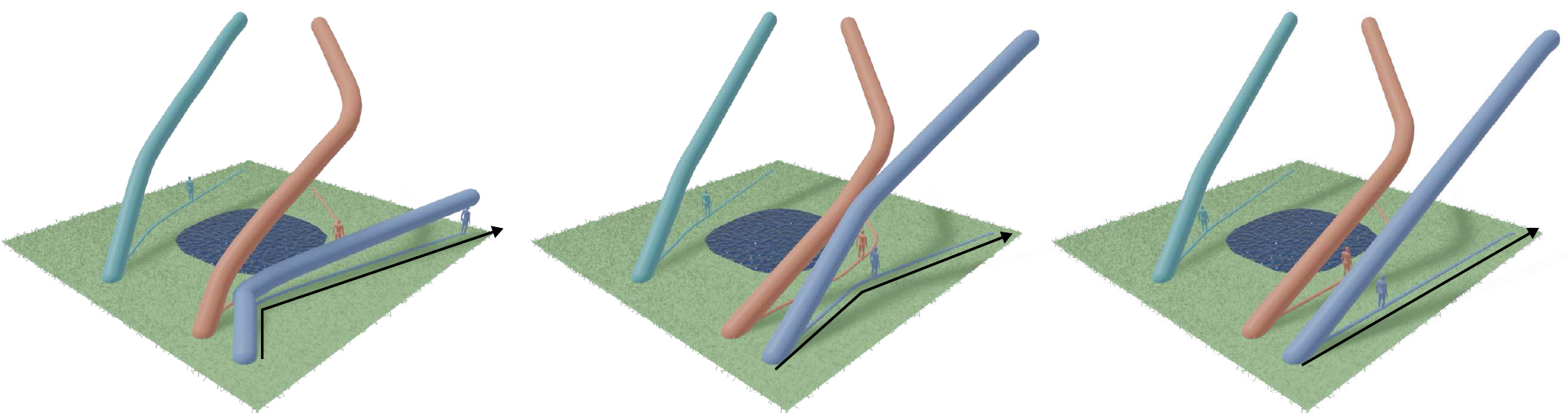}
\caption{Left: An agent delivers a message to another agent and rushes to his goal position. Middle: Two agents meet up briefly before walking to their separate destinations. Right: No contact between any agents.} 
\label{fig:pondScenes}
\end{figure}

\begin{figure}[h]
\centering
  \begin{subfigure}[t]{0.45\columnwidth}
  \includegraphics[width=\columnwidth]{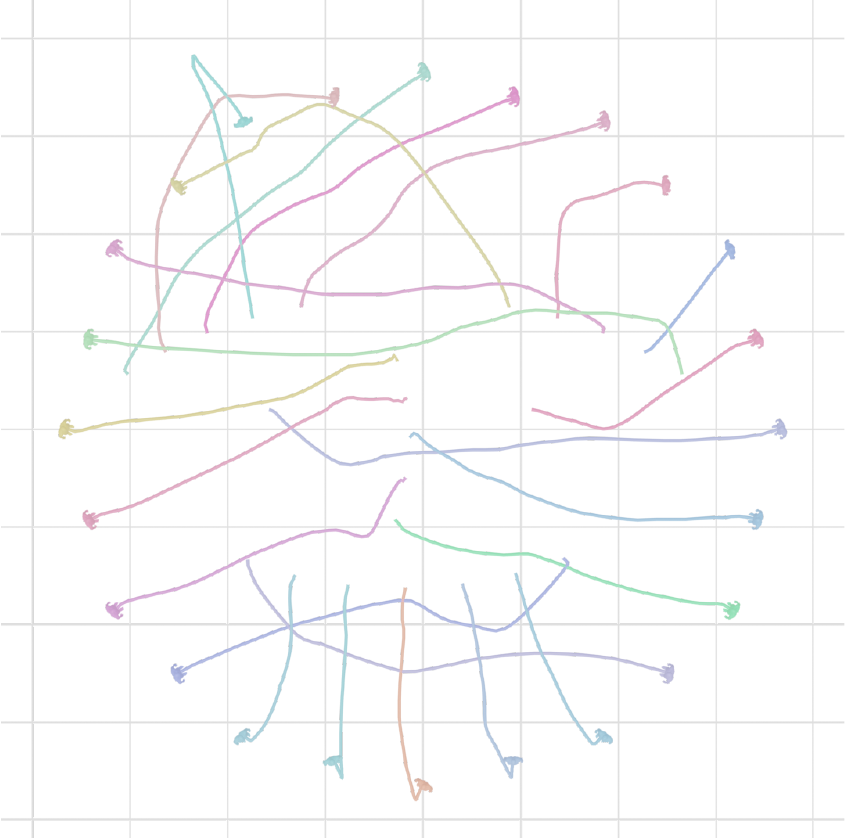}
  \caption{Pre-smile flash mob} 
  \label{fig:presmile}
  \end{subfigure}
  \begin{subfigure}[t]{0.45\columnwidth}
  \includegraphics[width=\columnwidth]{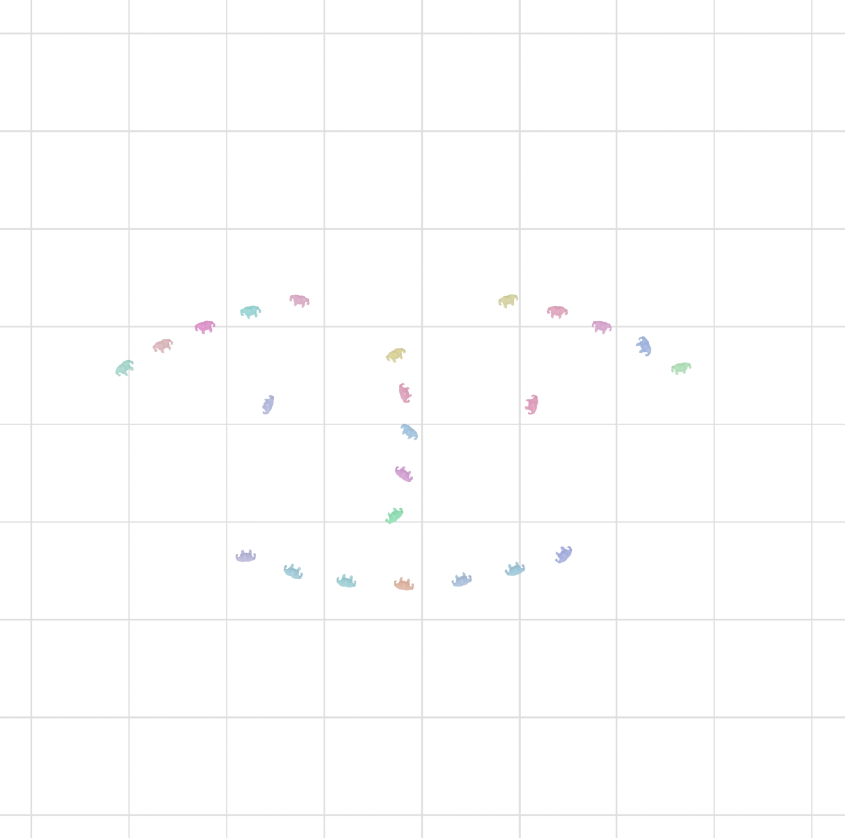}
  \caption{Post-smile} 
  \label{fig:postsmile}
  \end{subfigure}
  \caption{(a) group of people in a flash mob instructed to form a smily face. Agents follow unintuitive paths through space and time to maintain the fluidity of their motion. (b) a smily face created by controlling the motion of the mob.} 
  \label{fig:smilyFaceFlashMob}
\end{figure}

In~\autoref{fig:pondScenes} we show the flexibility and controllability of our method. We create a scenario where one agent must meet up with another agent to deliver a message and then arrive at his end goal faster than the other agents. In another scenario, we direct the agent to meet up with the second agent, yet arrive at his end goal at the same time as the other agents. Lastly, we also provide the simple scene where agents are ignorant of each other. The flash mob example in~\autoref{fig:smilyFaceFlashMob} shows how our method can be used to position agents intricately. The corresponding submission video shows how we are able to control the order of the placement of the agents in the flash mob as well through the preferred end time parameter. Our method allows careful control with no manual effort on the part of the user, thus making it useful for games or other industrial applications. 

And finally, for the sake of completeness we show~\autoref{fig:102Agents}, a large scene with 104 agents. We include this example to demonstrate that even though our method is designed for small to medium sized scenes, it can work on simple scenes with larger numbers of agents with asymmetric interactions and still scales linearly in time.

\subsection{Timings}
~\autoref{fig:timings} provides three plots, each with different scaling information for different usage scenarios. \vm{The first plot measure the average time per solver function iteration for an increasing number of agents. Each agent trajectory has 100 nodes (300 DOFs) and the performance is linear with broad-phase collision detection enabled. Our second plot measure the average time per iteration for an increasing number of DOFS for an 8 agent circle. The number of nodes starts off at 30 and goes all the way to 180 nodes per agent. Performance is very nearly linear with broad phase collision detection enabled. Without a broad phase, performance is quadratic.} Something to note is that increasing the number of DOFs per agent provides no additional benefit to the quality of the simulation since the actual physics of the space-time rods are not the end result of our method. As long as agents trajectory curves have enough DOFs to traverse the environment smoothly, the end results will be good. Our third plot measures the time per iteration for an increasing complexity in the environment mesh on the corn maze (~\autoref{fig:cornMaze}) example with three agents and 200 trajectory nodes per agent. Again, performances is nearly linear.

As mentioned before there is no gold-standard crowd simulation algorithm since each scenario is so unique and intricate. Rather than focus on timing performance for large crowds, our method focuses on solving previously overlooked path planning scenarios with asymmetric interactions for small to medium sized groups in highly constrained environments. \vm{Our asymptotics show linear to near-linear performance, but there is plenty of room for improvement in wall-clock-times through optimization and parallelization.}

\section{Conclusion and Future Work}
In this paper we show that modeling the motion of agents through space and time using a 3D curve resolves a number of difficulties with multi-agent path planning. 
Assigning physical characteristics such as a radius and mass to the agent's trajectory curve lets us intuitively simulate asymmetric interactions between agents. 
Our method outputs agent paths that are smooth, can navigate through highly constrained environments, and are parameterized by intuitive controls. 

In the future we hope to improve our agent model to include limited environmental perception (rather than the omniscience our agents enjoy currently) as well as additional dynamics. \vm{As mentioned before, performance depends on two factors: number of agents and their temporal support. While we can currently push along these axes independently, a future work is to be able to push along both axes together, i.e. handle complex environments with large numbers of agents with intricate motion. We also hope to extend the method in order to handle scenes with conflicting and changing goals, e.g. a predator-prey scenario. We would also like to extend our space-time approach to fully 3D environments which requires performing our optimization in four dimensions.
Finally, while our focus is on robustness in path planning, there is still further room for improvement in performance.} We are excited to explore fast space-time multi-agent path planning to scale our approach to the large dense crowds that continuum approaches excel at, while maintaining our unique advantages.

\printbibliography   
\end{document}